\documentclass[final]{cvpr}

\usepackage{times}
\usepackage{epsfig}
\usepackage{graphicx}
\usepackage{amsmath}
\usepackage{amssymb}
\usepackage{url}
\usepackage{comment}
\usepackage{multirow}
\usepackage[table,xcdraw]{xcolor}
\usepackage{color,colortbl}
\usepackage{subcaption}
\usepackage[labelformat=simple]{subcaption}

\usepackage[pagebackref=true,breaklinks=true,colorlinks,bookmarks=false]{hyperref}

\usepackage[hyphens]{url}
\hypersetup{breaklinks=true}
\usepackage{cite}

\def\cvprPaperID{11} % *** Enter the CVPR Paper ID here
\def\confYear{CVPR 2021}

\begin{document}

%%%%%%%%% TITLE
\title{Discovering Multi-Hardware Mobile Models via Architecture Search}

\author{Grace Chu \and Okan Arikan \and  Gabriel Bender \and Weijun Wang \and Achille Brighton \and Pieter-Jan Kindermans \and Hanxiao Liu \and Berkin Akin \and Suyog Gupta \and  Andrew Howard \and
Google LLC\\
{\tt\small \{cxy, okana, gbender, weijunw, aib, pikinder, hanxiaol, bakin, suyoggupta, howarda\}@google.com}
% For a paper whose authors are all at the same institution,
% omit the following lines up until the closing ``}''.
% Additional authors and addresses can be added with ``\and'',
% just like the second author.
% To save space, use either the email address or home page, not both
}

\maketitle

%%%%%%%%% ABSTRACT
\begin{abstract}

Hardware-aware neural architecture designs have been predominantly focusing on optimizing model performance on single hardware and model development complexity, where another important factor, model deployment complexity, has been largely ignored. In this paper, we argue that, for applications that may be deployed on multiple hardware, having different single-hardware models across the deployed hardware makes it hard to guarantee consistent outputs across hardware and duplicates engineering work for debugging and fixing. To minimize such deployment cost, we propose an alternative solution, multi-hardware models, where a single architecture is developed for multiple hardware. With thoughtful search space design and incorporating the proposed multi-hardware metrics in neural architecture search, we discover multi-hardware models that give state-of-the-art (SoTA) performance across multiple hardware in both average and worse case scenarios. For performance on individual hardware, the single multi-hardware model yields similar or better results than SoTA performance on accelerators like GPU, DSP and EdgeTPU which was achieved by different models, while having similar performance with MobilenetV3 Large Minimalistic model on mobile CPU. \footnote{Multi-hardware models (Multi-AVG and Multi-MAX) are available at {\scriptsize \url{https://github.com/google-research/google-research/tree/master/tunas}} and {\scriptsize \url{https://github.com/tensorflow/models/blob/master/official/vision/beta/modeling/backbones/mobilenet.py}}}

\end{abstract}

\begin{figure*}
\vspace{-2mm}
    \centering
    \includegraphics[width=1.7\columnwidth]{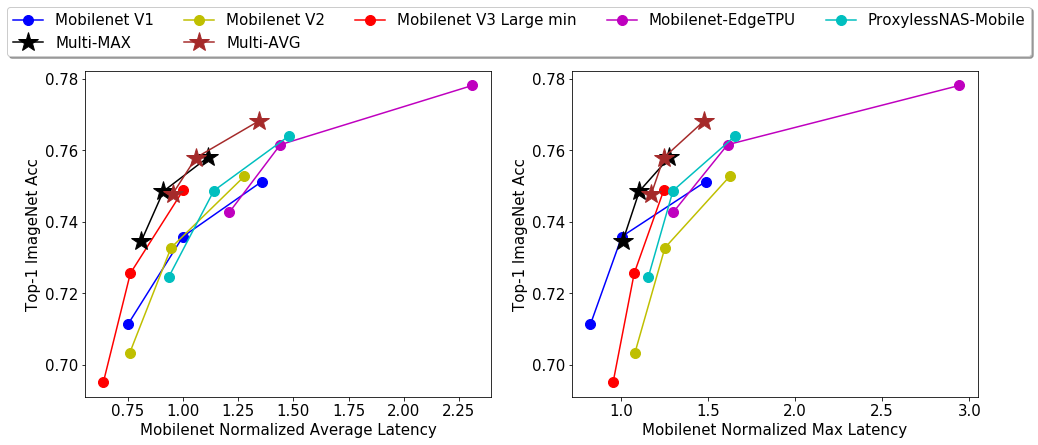}
    \vspace{-3mm}
    \caption{ImageNet test accuracy v.s. average latency (Mobilenet Normalized Average Latency in the left figure) and worst case latency (Mobilenet Normalized Max Latency in the right figure) over 5 hardware in Pixel4: CPU float, CPU uint8, GPU, DSP and EdgeTPU. See Section \ref{subsec:mobilenet_norm_metric} for details of how the average and worst case latency on x-axis are calculated. Multi-MAX and Multi-AVG are discovered multi-hardware models which yield the SoTA accuracy-latency tradeoff w.r.t. both average and worst case performance.}
    \label{fig:p4_overall_perform}
\vspace{-2mm}
\end{figure*}

%%%%%%%%% BODY TEXT
%\vspace{-2mm}
\section{Introduction}
\label{sec:intro}
%\vspace{-1mm}

Developing efficient on-device neural networks has become an important topic in computer vision with many real-world applications. Having models that can be fully deployed on device not only enables fast, real-time results, but also avoids exposing personal data to public servers.

Given the resource constraints of a portable device, such as latency, energy and memory footprint, on-device models need to be fast and small. While the number of multiply-and-add operations (MAdds) and the number of parameters have been widely used to optimize efficient models \cite{squeezenet,squeezenext,transfer_arch_learning,efficientnet,autoslim}, recent research has shown that improvements on theoretical MAdds or number of parameters do not always translate into better latency on real hardware, and can actually be counterproductive in some cases \cite{netadapt,mnas}. Thus, optimizing directly on latency measurements becomes important when we want to find a fast model on device \cite{mobilenet_v3,tunas}.

Unlike MAdds or the number of parameters, latency is highly dependent on hardware (and its associated software). A neural network optimized for a specific hardware platform may perform sub-optimally on a different one in terms of inference efficiency. Therefore, different models have been developed to achieve the best performance on each individual hardware \cite{fbnet, chamnet, gpu_model, once_for_all}.

However, for application developers who want to deploy their application on multiple hardware, using different single-hardware models for each hardware introduces deployment overhead from multiple aspects. For example, one needs to tune multiple models and dependent components in the system to guarantee consistent application outputs across hardware. In addition, duplicated debugging and fixing work are needed for any issue or update of the application and the complexity increases linearly with the number of deployed hardware.

To solve these problems, we propose multi-hardware models, where a single model is developed by optimizing on multiple hardware. It minimizes model deployment cost while still performs reasonably good on each targeted hardware. The contributions of this paper can be summarized as follows.

%it brings about challenges for application developers who try to decide which neural networks to adopt, especially when they want to deploy the application on multiple hardware and expect good performance on each of them.

%Comparing with using different models for different hardware, having a single model for multiple hardware reduces training, inference, storage and maintenance overhead for applications that may be deployed on multiple hardware. Specifically, application developers do not need to maintain multiple models for the same application across different hardware, which simplifies the system and makes it easy for quality control and bug tracking. 

%For above reasons, this paper aims to provide a single model solution for multiple hardware. The contributions can be summarized as follows.

\vspace{-3mm}
\begin{itemize}
\setlength\itemsep{-0.2em}
    \item It is the first work exploring the feasibility of multi-hardware models.
    \item It proposes a complementary search method that can be used by any existing neural architecture search (NAS) algorithms to find multi-hardware models.
    \item It proposes multi-hardware metrics to evaluate overall efficiency among multiple hardware.
    \item The discovered multi-hardware models via proposed method achieve SOTA performance across targeted hardware and can be generalized to other hardware as well.
\end{itemize}

\section{Related Work}
\label{sec:related_work}
%\vspace{-1mm}

\noindent {\bf Hardware-aware neural architecture designs} have been a popular research area in recent years. NetAdapt \cite{netadapt} uses empirical latency tables of the target hardware to greedily adapt a model to its highest accuracy under a target latency constraint. MnasNet \cite{mnas} also uses latency tables, but applies reinforcement learning to do hardware-aware architecture search. FBNet \cite{fbnet} and ChamNet \cite{chamnet} find the best architecture for targeted hardware by incorporating latency table and resource predictive models in architecture search respectively. MoGA \cite{gpu_model} optimizes a model for GPU. Once-for-all \cite{once_for_all} proposes a pre-trained super-model where different sub-models can be extracted for different hardware.

\noindent {\bf Neural architecture search (NAS)} has been widely used in hardware-aware architecture designs as the unpredictable hardware performance of a model makes it challenging to optimize models by hand \cite{nas,transfer_arch_learning,pham2018enas,mnas,tunas,hw_nas_benchmark,hournas}. This technique uses reinforcement learning \cite{nas}, evolutionary search \cite{evolutionary_search}, differentiable search \cite{diff_search,diff_search2} or other algorithms \cite{nas_other_algorithm} to find the best neural architecture according to a predefined reward function which incorporates both model performance and hardware efficiency \cite{hw_nas_benchmark}.

\section{Why Multi-Hardware Models}
\label{sec:why-multi-hardware}
%\vspace{-1mm}

Existing hardware-aware architecture designs have primarily focused on single-hardware models which aim to deliver the best model for each single hardware. With the recent advance of NAS \cite{hournas} and smart training algorithms \cite{once_for_all}, the cost of architecture search and model training have been significantly reduced. However, the outcome of $N$ models for $N$ hardware introduces much overhead for application developers who want to deploy the application on multiple hardware.

First overhead comes from the component level tuning when deploying the model. For example, when using the classification results to suggest a user action, like scanning text in the image, or blurring background of a portrait photo, one needs to tune a score threshold to determine when to give that suggestion. Then, if $N$ models are used for $N$ hardware, multiple score thresholds need to be tuned to ensure consistent performance across hardware.

Moreover, having different models on different hardware for the same application makes it hard to keep consistent model performance across hardware. Besides the overhead of tuning multiple models to have the same accuracy, it is almost impossible to make sure that all models give the exactly same output for every image it received. 

In addition, when unexpected performance occurs, such as the model false triggers or misses certain images, one needs to first determine whether it is a universal issue, or it only happens on some hardware as different models are used. Every debugging and fix step needs to be done $N$ times for each different model used for this application.

Last but the not the least, having a single model for multiple hardware on the same device, like CPU and EdgeTPU on Pixel4, enables seamless transition of workload from one hardware onto another other at runtime. In addition, only one model needs to be stored in this case.

In summary, multi-hardware model is a solution to minimize deployment complexity, a factor that has been largely ignored in developing single-hardware models. We will show in this paper that, despite the challenges, with proper balance among hardware in search space design and metric definition, multi-hardware models can still yield decent results across a wide range of hardware.

\section{Challenges of Multi-Hardware Models}
\label{sec:challenges}

\begin{figure}[t]
    \centering
    \includegraphics[width=85mm]{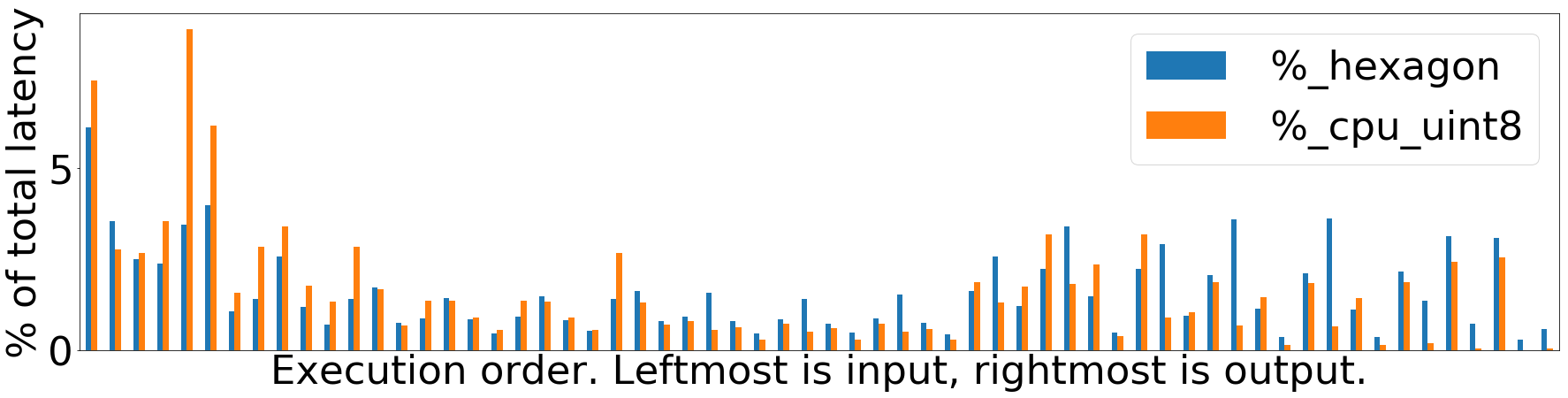}
    \vspace{-6mm}
    \caption{Per layer profiling of MobileNetV3 minimalistic on Pixel4 CPU uint8 and DSP (Hexagon). The leftmost is the input layer while the output layer is on the right.}
    \label{fig:diff_profile}
\vspace{-2mm}
\end{figure}

\noindent{\bf Diverse design preferences}: Due to the unique design of each hardware, their specialties are usually different, which may yield different directions of optimization. To demonstrate this, we take the MobilenetV3 Large minimalistic (min) model \cite{mobilenet_github} and run per layer profiling on CPU (uint8) and DSP (Qualcomm 855 Hexagon) to get the latency percentage of each layer over the whole model (Figure \ref{fig:diff_profile}). On CPU, a larger fraction of the model's total latency comes from the earlier layers of the network, while on DSP, a larger fraction of the total latency comes from the later layers. Therefore, when optimizing on CPU, one may focus mainly on the early layers, while later layers may gain more attention when optimizing for DSP.

\noindent {\bf Different supported operations}: While new operations and model blocks have been proposing to improve accuracy and latency trade-offs \cite{se,shufflenet,mobilenet,mobilenet_v3}, not all of them are equally efficient on different hardware. For example, the depthwise separable convolution that was proposed in MobileNet \cite{mobilenet} to replace the regular convolution reduces MAdds and makes model inference more efficient on CPU. However, a decrease in MAdds does not always lead to a decrease in on-device latency, especially for accelerators which have been optimized specifically to handle large number of computations as long as they follow certain pattern \cite{eyeriss}. For example, \cite{tpu_model} indicates that EdgeTPU favors regular convolution over depthwise separable convolution in certain layers of the model as the former can utilize the hardware resources better and gives better latency-accuracy trade-offs. This makes it hard to manually decide what operation to use at which layer if we want to have a single model that works well on both CPU and EdgeTPU.

\begin{figure}[t]
    \centering
    \includegraphics[width=85mm]{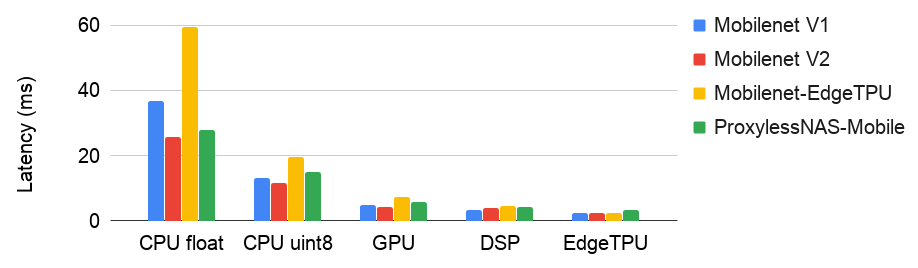}
    \vspace{-8mm}
    \caption{\small Latency of different models on different hardware in Pixel4 phone.}
    \label{fig:diff_ratios}
\vspace{-2mm}
\end{figure}

\noindent {\bf Diverse latency relationship}: It is well known that a model has different latencies when running on different hardware, but is the relationship similar for all models? That is, if a model runs 2$\times$ faster than another on one hardware, 1) will it still run faster on another hardware? 2) if faster, will it still be 2$\times$ faster? To answer this questions, we take four mobile models, MobileNetV1 \cite{mobilenet}, MobileNetV2 \cite{mobilenet_v2}, MobileNet EdgeTPU \cite{tpu_model}, ProxylessNAS mobile \cite{proxylessnas}, and benchmark them on different hardware to see whether the latency ratio among them are the same. Figure \ref{fig:diff_ratios} shows the results, where the latency ratio among different hardware are obviously different for each model. Furthermore, while EdgeTPU runs MobileNet-EdgeTPU model faster than ProxylessNAS-Mobile, CPU executes ProxylessNAS-Mobile faster than MobileNet-EdgeTPU.

\section{Problem Formulation}
\label{sec:prob_form}

Due to the challenges listed in previous section, manually handcrafting a single model to accommodate the traits of multiple hardware is very difficult. Instead, we leverage neural architecture search to find multi-hardware models in this paper, where a multi-hardware search space is proposed to be compatible to all examined hardware, and two metrics are introduced to compare models under multi-hardware environment.

\subsection{Multi-Hardware Search Space}
\label{subsec:multi_hw_search_space}

Let $\mathcal{H} = \{H_1, H_2, ..., H_N\}$ be the set of hardware we want to optimize for. For $0<i\leq N$, $S_i$ denotes the set of neural network architectures that $H_i$ can support, i.e., the entire network can fully run on this hardware without falling back to another slower hardware. Then, a multi-hardware search space, denoted as $S^\mathcal{H}$, is a set of neural network architectures that belongs to the intersection of supported architectures of the set of examined hardware. Mathematically,
\vspace{-1mm}
\begin{equation}
    S^\mathcal{H} \subseteq S_1 \cap S_2 \cap \dots \cap S_N.
\vspace{-1mm}
\end{equation}
Note that, we allow the multi-hardware search space to be a subset of instead of equal to the intersection of all supported architectures, by taking into account the practical size limit for efficient architecture search.

\subsection{Multi-Hardware Metrics}
\label{subsec:multi_hw_metrics}

In order to find a single model optimized for multiple hardware, we need metrics to determine what is a better model. Without loss of generality, we examine models' accuracy and latency to compare different models, as the Pareto optimal on these two metrics has been broadly used in single-hardware architecture optimizations \cite{mnas,tunas,mobilenet}. Specifically, model $a$ is better than model $b$ in single-hardware optimization iff
\vspace{-2mm}
\begin{equation}
    L_a < L_b \text{ when } A_a = A_b,
\vspace{-2mm}
\end{equation}
where $(L_a, A_a)$ and $(L_b, A_b)$ are the (latency, accuracy) measurements of model $a$ and $b$ on the examined single hardware, respectively.

When considering multi-hardware optimization, the biggest challenge is how to compare models given their latency measurements on various hardware. Let $\mathcal{L}_a = \{L_{a,1}, L_{a,2}, \dots, L_{a,N}\}$ denote the latency of model $a$ on $\mathcal{H}$, similarly for model $b$. We need some overall metric function $f^\mathcal{H}(\cdot)$ such that, if
\vspace{-2mm}
\begin{equation}
    f^\mathcal{H}(\mathcal{L}_a) < f^\mathcal{H}(\mathcal{L}_b) \text{ when } A_a = A_b,
\vspace{-2mm}
\end{equation}
we say that model $a$ is better than model $b$.

As shown in Section \ref{sec:challenges}, latency on different hardware may have different scales, thus $L_{a,i}$ needs to be normalized before any calculation. In this paper, we propose two intuitive metrics to measure the average and worst case performance of a model on multiple hardware. Specifically, the normalized average latency over $\mathcal{H}$ is defined as
\vspace{-2mm}
\begin{equation}
\label{eq:avg}
    f^\mathcal{H}_{avg}(\mathcal{L}_a) \triangleq \frac{1}{N}\sum_{i=1}^{N} \frac{L_{a,i}}{C_i},
\vspace{-2mm}
\end{equation}
and the normalized max latency over $\mathcal{H}$ is defined as
\vspace{-2mm}
\begin{equation}
\label{eq:max}
    f^\mathcal{H}_{max}(\mathcal{L}_a) \triangleq \max_i \bigg( \frac{L_{a,i}}{C_i} \bigg),
\vspace{-2mm}
\end{equation}
where $\mathcal{C} = \{C_1, C_2,\dots, C_N\}$ are normalization factors. While there are many ways of choosing $\mathcal{C}$, we discuss two common cases as follows.

\vspace{-2mm}
\begin{enumerate}
\setlength\itemsep{-0.2em}
    \item $\mathcal{C}$ can be chosen as the latency of a reference model on $\mathcal{H}$ to represent the latency scaling relationship among hardware. In addition, if the normalized average latency of a model is 0.5, it implies that, on average, the model runs in half of the time of the reference model.
    \item On top of the natural latency scaling difference among hardware, one can further re-weight $\mathcal{C}$ with the importance of each hardware in $\mathcal{H}$. An extreme case would be to set all norm factors to be $\infty$ except one $C_1=1$. Then $f^\mathcal{H}(\mathcal{L}_a)=L_{a,1}$, which regresses the problem to a single hardware optimization.
\end{enumerate}
\vspace{-2mm}

\textit{Remark}: This paper mainly focuses on mobile models as cross device application is the most common use case of multi-hardware models. However, the methodology introduced here can be easily generalized to discover multi-hardware server sized models when needed.

\section{Case Study}
\label{sec:case_study}

With the essential concepts defined above, we use an on-device case study to demonstrate how to find multi-hardware models via architecture search. Here, we consider five hardware inside a Pixel4 phone: CPU float, CPU uint8, GPU (Qualcomm Adreno 640), DSP (Qualcomm Snapdragon 855), EdgeTPU (Google). We choose this set of hardware because
\vspace{-2mm}
\begin{itemize}
\setlength\itemsep{-0.2em}
    \item it covers various types of hardware from different manufacturers for mobile;
    \item the obtained multi-hardware model is useful in application: Because it performs well on all hardware on Pixel4, it can be used as a default model for any application deployed on Pixel4 regardless of its particular inference hardware.
\end{itemize}
\vspace{-2mm}
We use TuNAS \cite{tunas} as the NAS infrastructure to use in this paper, while the proposed method can be applied on many existing NAS algorithms to get multi-hardware models.

\subsection{Multi-Hardware Search Space on Device}
\label{subsec:multi_hw_search_space_device}

In order to optimize for multiple hardware, multi-hardware search space needs to be both exclusive enough so that each searched operation is supported by all examined hardware, and inclusive enough so that it searches over a variety of effective (and supportive) operations for each examined hardware.

\vspace{-2mm}
\begin{itemize}
\setlength\itemsep{-0.2em}
    \item We center the search space at MobilenetV3 Large model's architecture as it is one of the SoTA mobile models.
    \item We remove squeeze and excite (SE) and h-swish because they are not supported in EdgeTPU.
    \item Filter sizes are adjusted to be integer multiples of 32 due to a preference of DSP \cite{snapdragon_limit}.
    \item Similar to TuNAS MobilenetV3 Large search space \cite{tunas}, we search over the number of repeated blocks per stage from \{1, 2, 3, 4\}, the expansion ratio from \{1, 2, 3, 4, 5, 6\}, and the input/output filter size ratios from \{0.5, 0.625, 0.75, 1.0, 1.25, 1.5, 2.0\}.
    \item We do not search the input/output filter sizes in the model head because in early experiments we found that, the RL controller was biased towards using large numbers of filters in the model head which blew up the model size but with marginal accuracy improvement.
    \item Each block can choose either regular inverted bottleneck or fused inverted bottleneck, which replaces expansion 1x1 convolution (conv) and depthwise conv with a single regular conv, as it has been shown to be effective for EdgeTPU \cite{edgetpu}.
    \item Convolution kernels can choose from 3x3 and 5x5 because bigger kernels than 5x5 are not widely supported by DSPs.
\end{itemize}

\subsection{Mobilenet Normalized Avg and Max Metrics}
\label{subsec:mobilenet_norm_metric}

Given we are optimizing for mobile models, we choose to use MobilenetV1 as the reference model to calculate the overall metrics, i.e., use its latency on the examined hardware as the normalization factors $\mathcal{C}$ in equation (\ref{eq:avg}) and (\ref{eq:max}). More specific reasons to choose this reference model are:
\vspace{-2mm}
\begin{itemize}
\setlength\itemsep{-0.2em}
    \item existed for a few years and yet still widely used;
    \item publicly available in multiple formats (TFLite, Caffe, etc.) for ML researchers to run benchmarks with;
    \item simple enough that can run on a wide variety of hardware.
\end{itemize}
\vspace{-2mm}
We do not have a particular preference on having better performance on some of the optimized hardware, so no extra re-weights were assigned to these normalization factors. 

The reward function used in architecture search needs to be adjusted with these new metrics too. In TuNAS, single hardware search maximizes the following reward function \cite{tunas}:
\vspace{-2mm}
\begin{equation}
\label{eq:single_r}
    r(\alpha) = A(\alpha) + \beta \Big|\frac{L(\alpha)}{L_0} - 1\Big|,
\vspace{-2mm}
\end{equation}
where $\alpha$ represents an architecture, $r(\cdot)$, $A(\cdot)$ and $L(\cdot)$ are reward, accuracy and latency of the architecture, respectively. $|\cdot|$ is absolute function. $L_0$ denotes latency target. $\beta < 0$ is an application-specific constant.

To search for multi-hardware models, the reward function becomes
\vspace{-2mm}
\begin{eqnarray}
\label{eq:avg_r}
    r_{avg}(\alpha) &=&  A(\alpha) + \beta |f^\mathcal{H}_{avg}(\alpha) - 1| \nonumber \\
    &=& A(\alpha) + \beta \Big|\frac{1}{N} \sum_{i=1}^N \frac{L_i(\alpha)}{C_i} - 1\Big|,
\vspace{-2mm}
\end{eqnarray}
when optimizing for average performance; and
\vspace{-2mm}
\begin{eqnarray}
\label{eq:max_r}
    r_{max}(\alpha) &=&  A(\alpha) + \beta |f^\mathcal{H}_{max}(\alpha) - 1| \nonumber \\
     &=& A(\alpha) + \beta \Big|\max_i \frac{L_i(\alpha)}{C_i} - 1\Big|,
\vspace{-2mm}
\end{eqnarray}
when optimizing for worst case performance. Note that when optimizing for average performance, the reward function implies a prior that the searched architecture should, on average, have latency close to that of the reference model MobilenetV1.

\section{Experiments}
\label{sec:experiments}

\subsection{Experimental Setup}

\noindent \textbf{Latency benchmarks}: In this paper, three phones with totally ten different hardware are used in either searching for the multi-hardware model or evaluating the model on unsearched hardware. The driver's versions for these phones are: Pixel4 uses QQ1B.200205.003; Pixel3 uses QQ1A.200205.002; MediaTek Dimensity 1000 5G uses QP1A.190711.020. 

The delegates used for accelerators are: GPU's latency is obtained from Jet delegate using OpenCL; DSP's latency is from Hexagon delegate which directly calls the Qualcomm's binary with less overhead than Android NNAPI; EdgeTPU's latency and APU's latency are obtained by using NNAPI delegate.

TF-Lite models with single-thread and batch size of 1 are used to get all benchmarking results. When getting CPU's latency, only the large cores were used. When benchmarking on CPU uint8, DSP, EdgeTPU and APU, where quantized models are needed, fake quantization is applied \cite{fake_quant}\footnote{In order to have consistency across multiple hardware, accuracy in this paper is always measured on the float model.}.

\begin{figure}
    \centering
    \begin{subfigure}[t]{0.19\columnwidth}
      \includegraphics[width=\columnwidth]{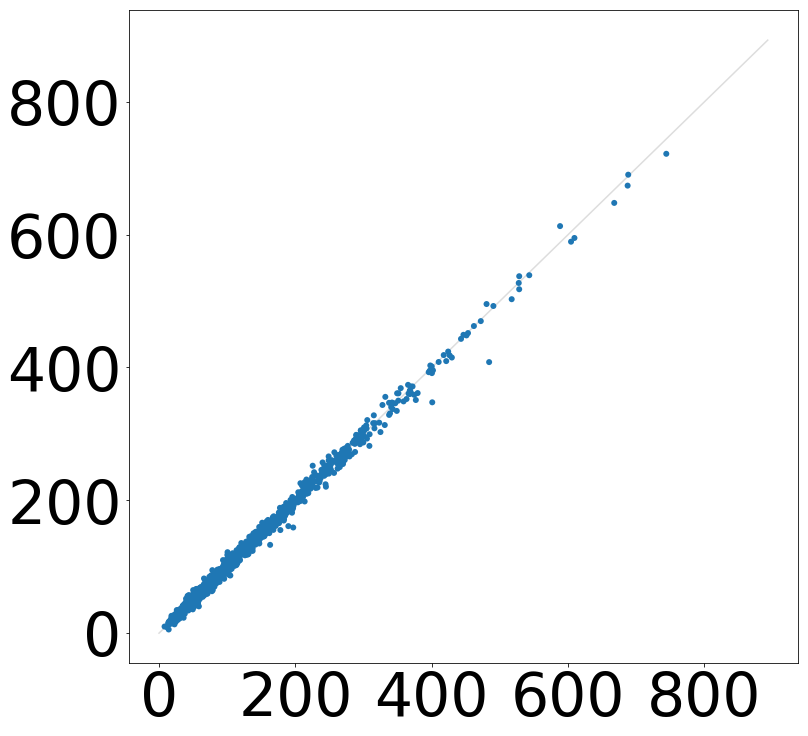}
      \caption{CPU float}
      \label{fig:calibration_a}
    \end{subfigure}
    \begin{subfigure}[t]{0.19\columnwidth}
      \includegraphics[width=\columnwidth]{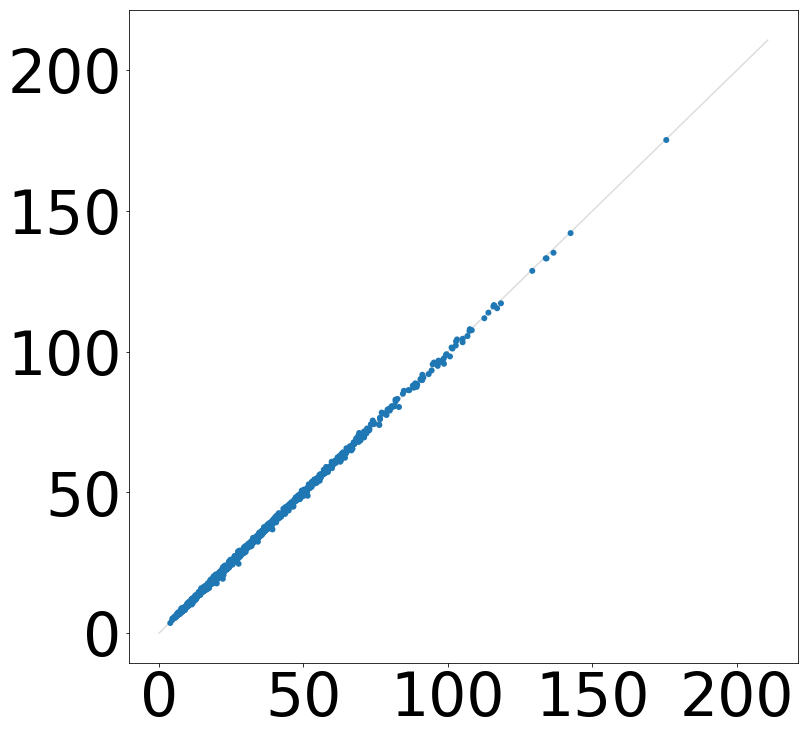}
      \caption{CPU uint8}
      \label{fig:calibration_b}
    \end{subfigure}
    \begin{subfigure}[t]{0.19\columnwidth}
      \includegraphics[width=\columnwidth]{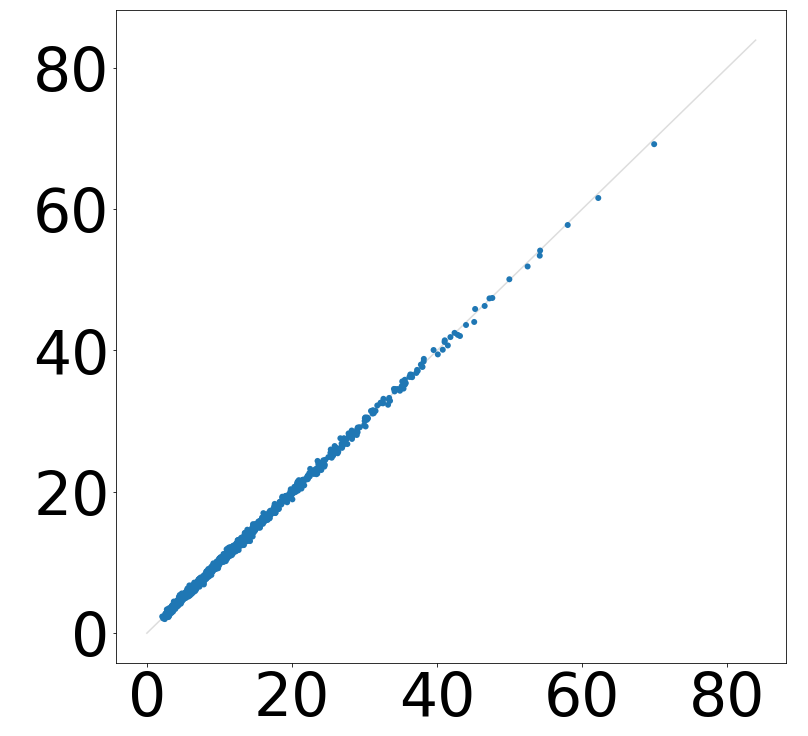}
      \caption{GPU}
      \label{fig:calibration_c}
    \end{subfigure}
    \begin{subfigure}[t]{0.19\columnwidth}
      \includegraphics[width=\columnwidth]{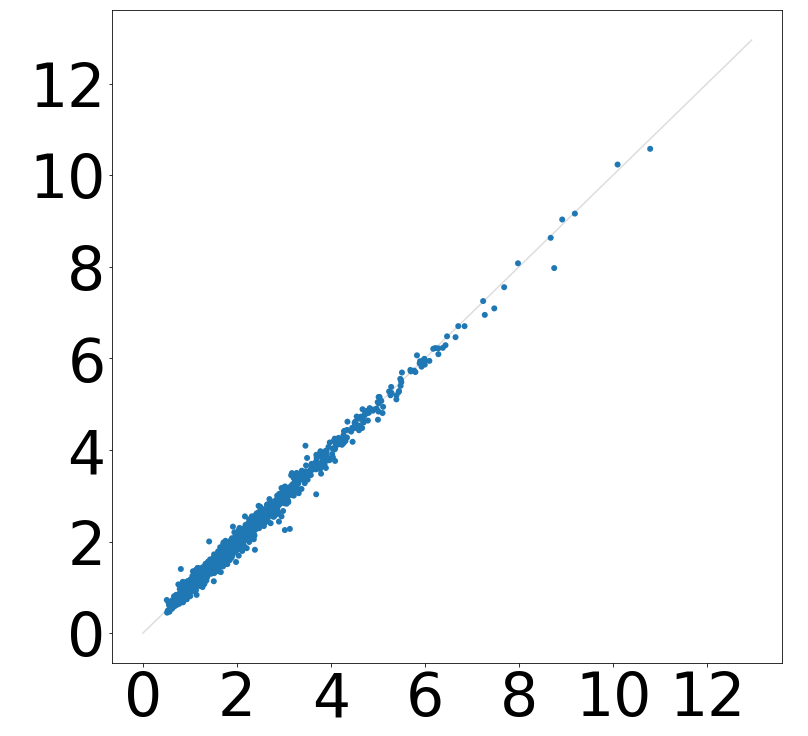}
      \caption{EdgeTPU}
      \label{fig:calibration_d}
    \end{subfigure}
    \begin{subfigure}[t]{0.19\columnwidth}
      \includegraphics[width=\columnwidth]{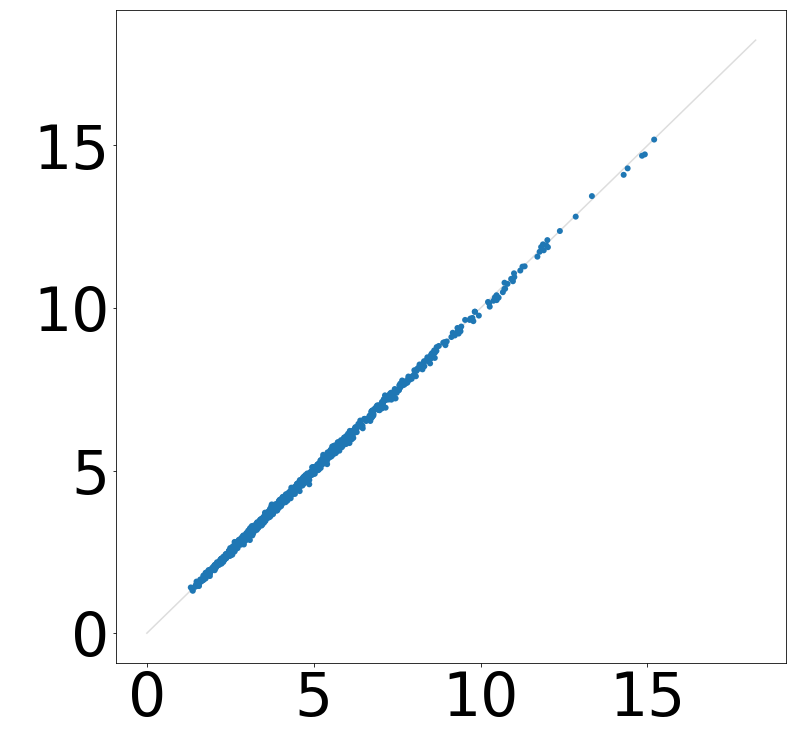}
      \caption{DSP}
      \label{fig:calibration_e}
    \end{subfigure}
    \vspace{-2mm}
    \caption{True cost/latency (x-axs) v.s. predicted cost/latency (y-axis) on unseen architectures from the search space for each of the trained cost models. Latency numbers are in millisecond unit.}
    \label{fig:calibration}
\vspace{-4mm}
\end{figure}

\begin{figure*}
\vspace{-3mm}
    \centering
    \includegraphics[width=1.7\columnwidth]{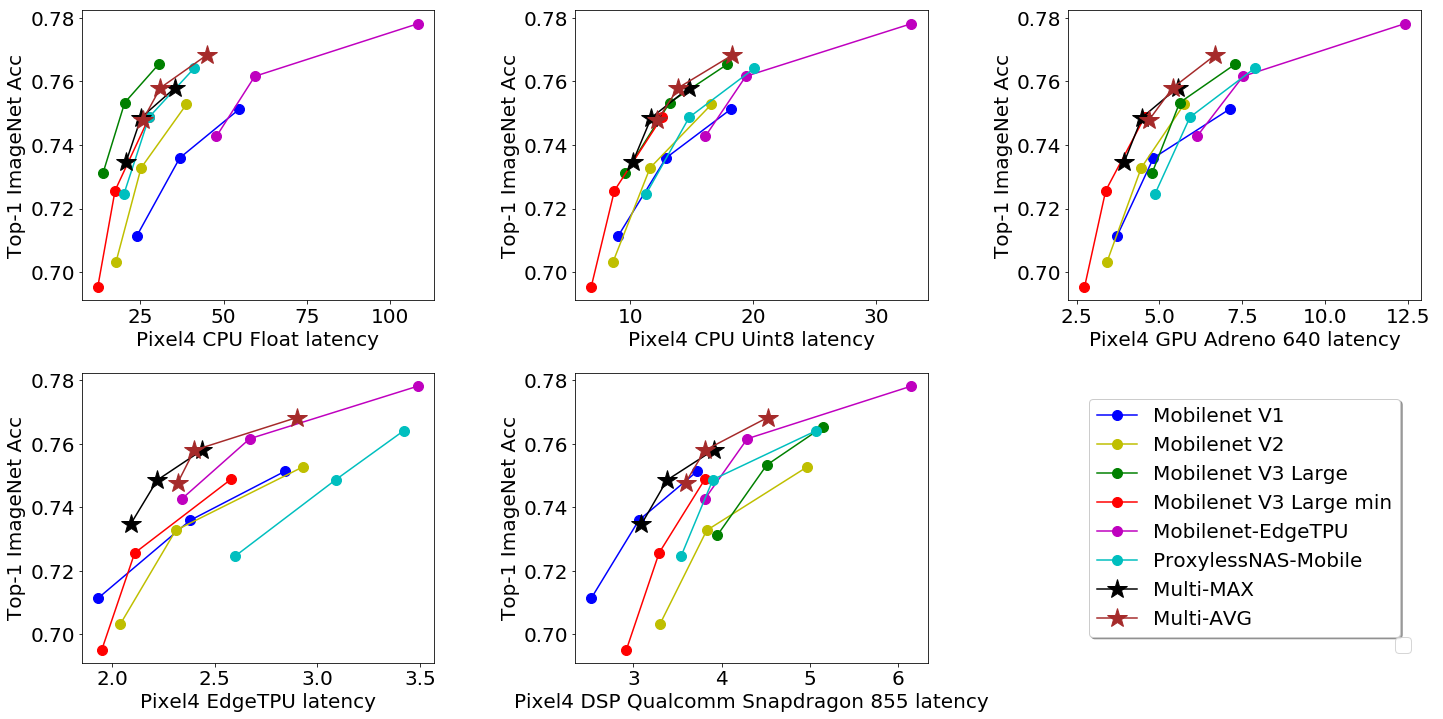}
    \vspace{-2mm}
    \caption{Accuracy-latency trade-offs of multi-hardware model v.s. baseline models on each optimized hardware. The horizontal axis is end-to-end real time latency benchmarks in milliseconds and the vertical axis is test accuracy.}
    \label{fig:per_hardware_pareto}
%\vspace{-5mm}
\end{figure*}

\begin{table*}[ht]
%\vspace{-7mm}
\caption{Performance of multi-hardware models comparing with baseline models on Pixel4 phone. `wm' is short for `width multiplier'. `$\times$' means that the model is not supported on that hardware. `MN-Norm' is the mobilenet normalized metrics proposed in Section \ref{subsec:mobilenet_norm_metric} where lower is better. Top-1 item within each column has been marked bold.}
\vspace{-2mm}
\centering
  \begin{tabular}{ l | c | c | c | c | c c c c c | c c }
    \hline
      & & Accu & Params & MAdds & \multicolumn{2}{c}{CPU} & & & & \multicolumn{2}{c}{MN-Norm}  \\
     \multirow{-2}{*}{Model} & \multirow{-2}{*}{wm} & (\%) & (M) & (M) & float & uint8 & \multirow{-2}{*}{GPU} & \multirow{-2}{*}{DSP} & \multirow{-2}{*}{EdgeTPU} & avg & max \\ 
      \hline
    MobilenetV1     &1.25&75.1    &6.25    & 883   & 54.7   & 18.2   & 7.12   & 3.72   & 2.84   & 1.36   & 1.49   \\ 
    MobilenetV2     &1.25&75.3    &5.01    & 487   & 38.8   & 16.6   & 5.74   & 4.97   & 2.93   & 1.28   & 1.62   \\
    MobilenetV3Large    &1.0 &75.3    &5.45    &\bf 217&\bf 20.3& 13.2   & 5.61   & 4.51   &$\times$&$\times$&$\times$\\
    MobilenetV3Large min &1.25&74.9    &5.73    & 346   & 27.7   & 12.6   & 4.56   & 3.81   & 2.58   & 1.00   & 1.25   \\
    ProxylessNAS-Mobile   &1.0 &74.9    &\bf 4.05& 321   & 27.6   & 14.8   & 5.92   & 3.90   & 3.09   & 1.14   & 1.30   \\
    \hline
    Mobilenet-EdgeTPU   &1.0 &\bf 76.2&\bf 4.05& 991   & 59.3   & 19.4   & 7.52   & 4.29   & 2.67   & 1.44   & 1.61   \\
    \hline
    Multi-AVG&1.0 &75.8    &4.91    & 433   & 31.0   & 13.9   & 5.40   & 3.81   & 2.40   & 1.06   & 1.25   \\
    Multi-MAX&1.0 &74.9    &4.39    & 349   & 25.2   &\bf 11.7&\bf 4.47&\bf 3.38&\bf 2.22&\bf 0.91&\bf 1.10\\
    \hline
  \end{tabular}
  \label{tb:table_comparison}
  \vspace{-3mm}
\end{table*}

\noindent \textbf{Cost models}: To save hardware communication cost in search, we pre-train linear cost models for target hardware so that the latency of any architecture sampled during search can be inferred from them. When training cost models, 9K pairs of (architecture, latency) data were used to train the cost model for each hardware, except for EdgeTPU where we used 20K to achieve the similar quality. Figure \ref{fig:calibration} shows that trained cost models have good correlations between predicted and true latency on unseen architectures.

\noindent \textbf{Architecture search and training}: We use ImageNet data \cite{imagenet} to search, train and evaluate. Input resolution is 224$\times$224 and ResNet data preprocessing is used. Cloud TPU v2-32 is used in both search and standalone model training, where per core batch size is 128. For standalone model training, we use the same hyper-parameters as in \cite{tunas}, where 0.25 is set as the dropout rate when training models for 360 epochs to get the test accuracy.

For architecture search, we increase the search length from what was used in \cite{tunas} as it shows some benefits when optimizing DSP and TPU. Specifically, 1) per core learning rate is halved from 0.0825 to 0.04125; 2) the warmup time where only shared model weights are trained without updating RL controller is increased from 25\% to 50\%; 3) we search for 360 epochs instead of 90 epochs.

\noindent \textbf{Baseline models}: To make fair comparisons, we re-implemented all baseline models in the training setup so that they use the same hyper-parameters with multi-hardware models. We have found similar trends as \cite{tunas} that the re-implemented MobilenetV1 and MobilenetV2 have higher accuracy numbers than the published ones in the original papers.

\begin{figure*}
\vspace{-3mm}
    \centering
    \includegraphics[width=1.7\columnwidth]{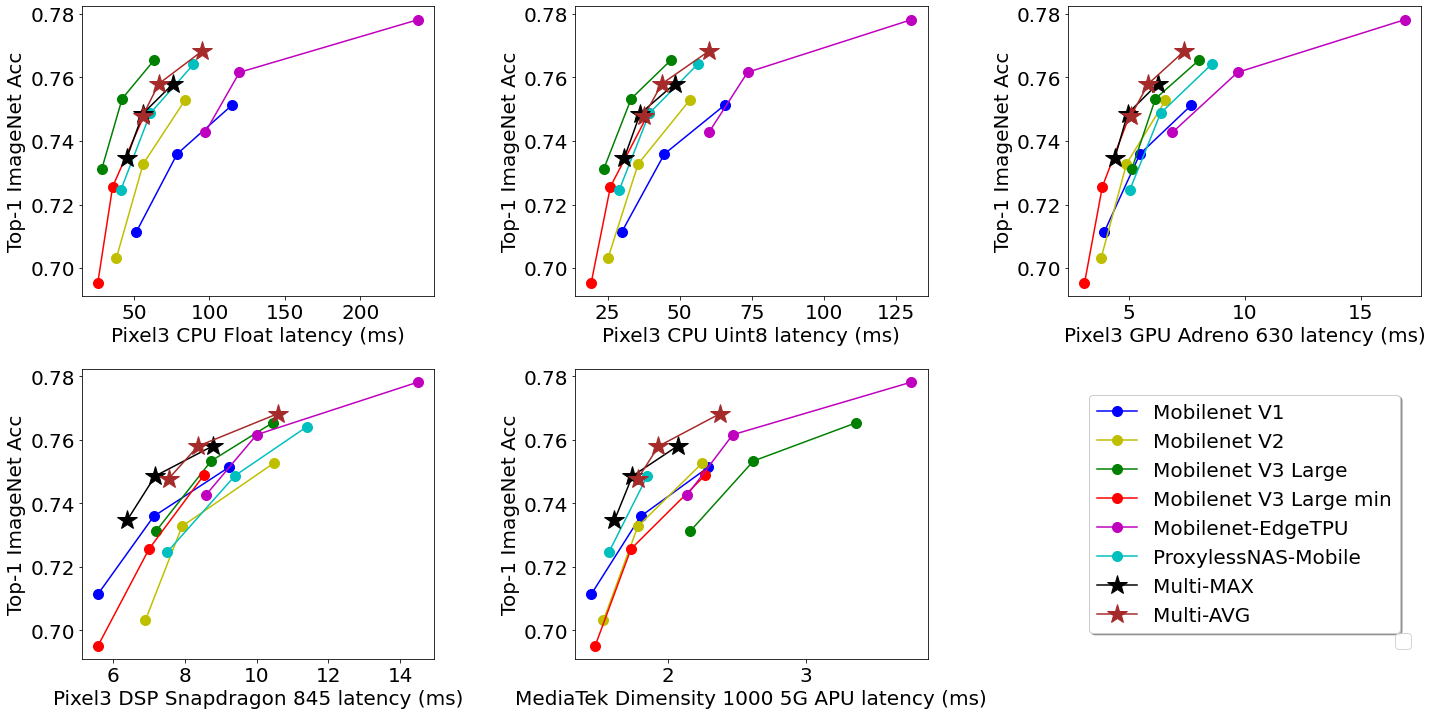}
    \vspace{-3mm}
    \setlength{\belowcaptionskip}{-8pt}
    \caption{Accuracy-latency trade-offs on unsearched hardware\protect\footnotemark, where the discovered Multi-MAX and Multi-AVG models are Pareto-optimal on all hardware except CPU.}
    \label{fig:unsearched}
\vspace{-3mm}
\end{figure*}

\subsection{Main Results}
\label{sec:search_results}

We conduct two multi-hardware architecture searches using TuNAS to find models that perform well on all hardware in Pixel4, regarding average performance and worst case performance, respectively. Both of them use the same multi-hardware search space proposed in Section \ref{subsec:multi_hw_search_space_device}. Reward functions used in these two searches are equation (\ref{eq:avg_r}) and (\ref{eq:max_r}) respectively. $\beta$ is set to -0.07.

Figure \ref{fig:per_hardware_pareto} shows the accuracy-latency pareto curves of the obtained multi-hardware models compared with (re-implemented) baseline models. `Multi-MAX' and `Multi-AVG' models are the results from searching over average metric and max metric, respectively. Each model has three points in the plot denoting the performance for width multiplier 0.75, 1 and 1.25.

On CPU float, except MobilenetV3 Large model, which is particularly optimized for CPU but not supported by EdgeTPU, multi-hardware models perform the best among all other baseline models. On the other four hardware platforms, multi-hardware models achieves SoTA trade-off between accuracy and latency. 

Specifically, on CPU uint8 and GPU, multi-hardware models perform similarly with MobilenetV3 Large min. However. they are much better than this baseline model on EdgeTPU and DSP. After updating MobilenetV1's accuracy with the better hyper-parameters, we found that this is the best baseline model on DSP. However, multi-hardware models still outperform MobilenetV1 when scaling up and the gap is much larger on other hardware. When comparing with Mobilenet-EdgeTPU, which is optimized particularly for EdgeTPU, multi-hardware models give better results on both EdgeTPU and other hardware.

Figure \ref{fig:p4_overall_perform} in the Introduction shows the overall performance where the normalization factors were taken as the latency of MobilenetV1 on examined hardware. As expected, the multi-hardware models are better than all baseline models in both average and worst case performance.

\begin{figure}
    \centering
      \includegraphics[width=0.49\columnwidth]{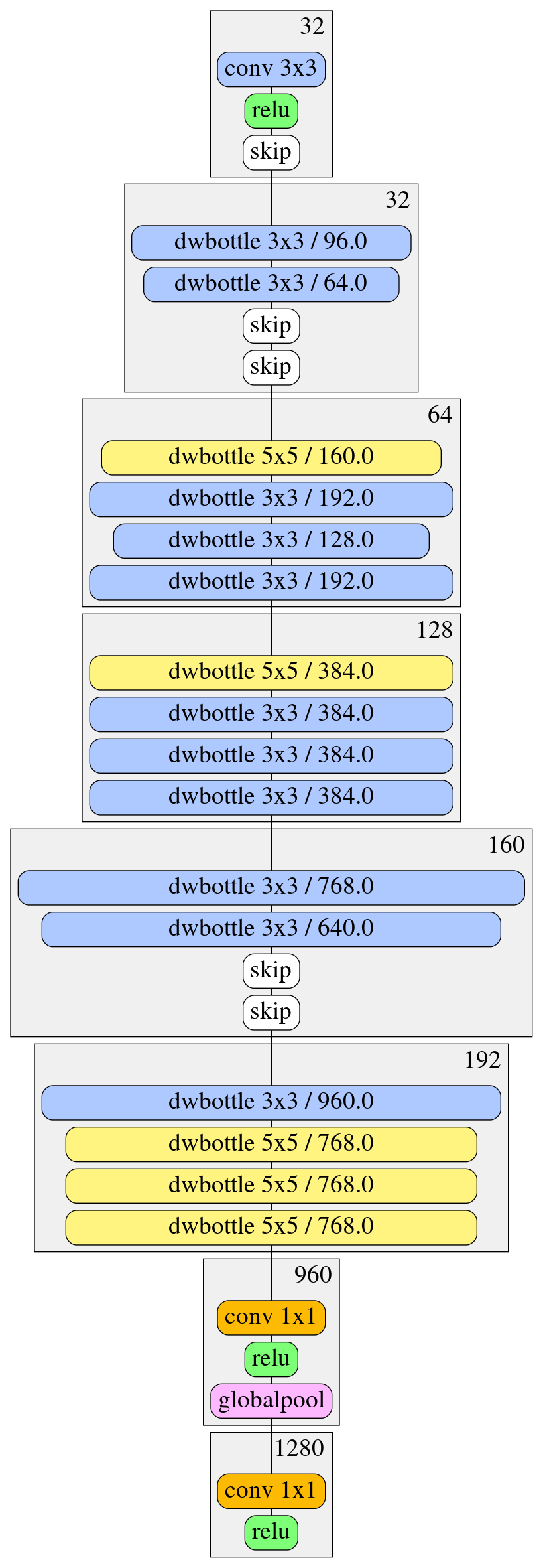}
      \includegraphics[width=0.49\columnwidth]{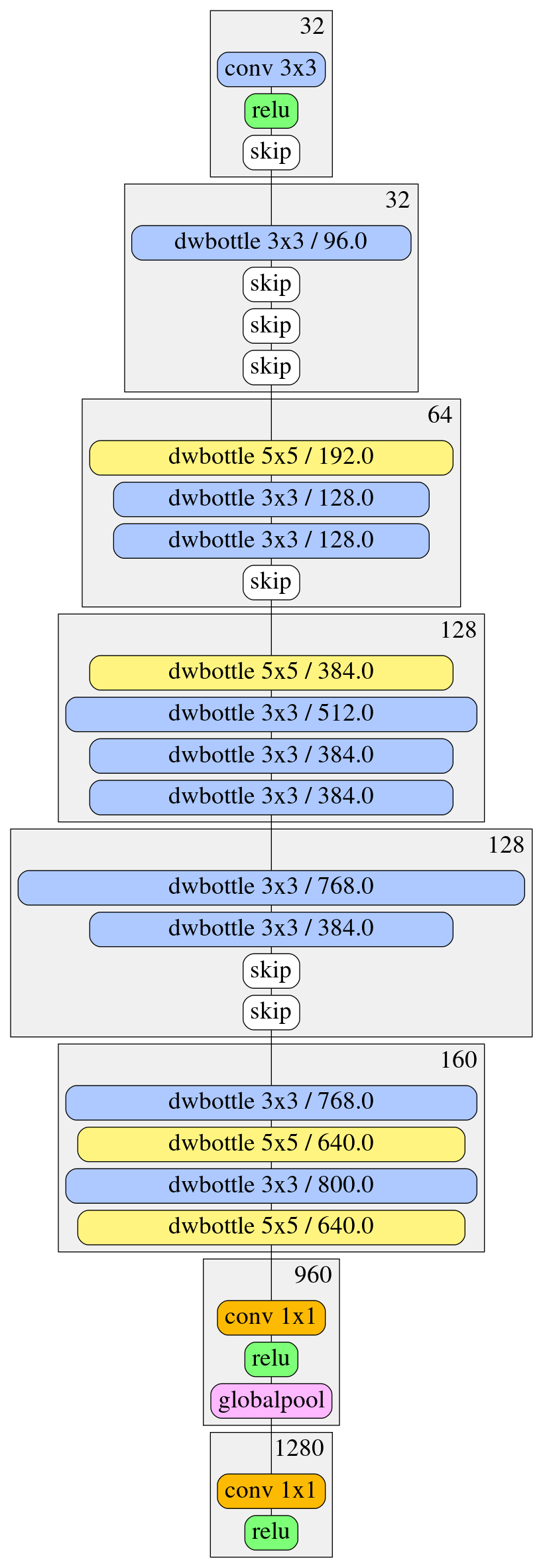}
      \vspace{-3mm}
    \caption{Model visualization of Multi-AVG (left) and Multi-MAX (right) models.}
    \label{fig:model}
    \vspace{-6mm}
\end{figure}

Numerically, we compare multi-hardware models with baseline models on similar accuracy range in Table \ref{tb:table_comparison}. Multi-MAX model runs the fastest on all examined hardware except on CPU float where it still ranks the second, while its accuracy is only 0.4\% lower than the second highest number in baseline models achieved by MobilenetV2 and MobilenetV3 Large. The top accuracy is achieved by Mobilenet-EdgeTPU, which is only 0.4\% higher than Multi-AVG model but its latency on CPU float is almost 2$\times$ slower and MAdds is 2.29$\times$ more. While MobilenetV3 Large achieves the best CPU float latency, it is not supported on EdgeTPU, and runs 18\% slower than Multi-AVG on DSP while also 0.5\% lower on accuracy. MobilenetV2 is 0.5\% lower on accuracy than Multi-AVG and runs also slower on all examined hardware: 25\% slower on CPU float and 30\% slower on DSP.

\footnotetext{\footnotesize{ProxylessNAS-Mobile only has two data points on MediaTek hardware as the model with width multiplier 1.25 is not fully supported by this hardware.}}

To show how multi-hardware models generalize on un-searched hardware, we evaluate their performance on various hardware of Pixel3 and MediaTek phones in Figure \ref{fig:unsearched}. Without optimizing for, multi-hardware models achieve SoTA performance on MediaTek accelerators. For Pixel3 hardware, multi-hardware models show similar trends as they are on Pixel4: yield the best results on Pixel3 GPU and DSP while being the second best results on CPU float. However, on CPU uint8, multi-hardware models do not give the best results on Pixel3 as they do for Pixel4. This is because Pixel3 CPU float and uint8 are similar hardware, while Pixel4 CPU uint8 has been particularly accelerated and performs much different from CPU float. The observation above demonstrates that one may only need to pick representative hardware to optimize, as the multi-hardware model will most likely have similar performance on closely related hardware, such as the same type of chips with different versions.

\begin{table*}[ht]
\vspace{-1mm}
\caption{Performance of single-hardware models. `Single-DSP' is the searched model only optimized for DSP. Top-1 item within each column is marked bold.}
\vspace{-3mm}
\centering
  \begin{tabular}{ l | c | c c c c c | c c }
    \hline
      & Accu & \multicolumn{2}{c}{CPU} & & & & \multicolumn{2}{c}{MN-Norm}  \\
     \multirow{-2}{*}{Model} & (\%) & float & uint8 & \multirow{-2}{*}{GPU} & \multirow{-2}{*}{DSP} & \multirow{-2}{*}{EdgeTPU} & avg & max \\ 
      \hline
    Single-CPU float                  &\bf 76.5& 39.6   & 18.0   & 6.23   & 4.52   & 3.32   & 1.33   & 1.48   \\ 
    \rowcolor{yellow} Single-CPU uint8&76.2    & 38.6   &\bf 13.9& 5.85   & 3.71   & 2.55   &\bf 1.13&\bf 1.21\\
    Single-GPU                        &76.0    &\bf 33.6& 15.6   &\bf 5.46& 4.10   &2.68    &1.15    &1.34   \\
    Single-DSP                        &76.3    & 46.7   & 15.6   & 6.88   &\bf 3.42&\bf 2.44& 1.21   & 1.43   \\
    Single-EdgeTPU                    &    76.0& 44.0   & 15.4   & 6.03   & 3.98   & 2.47   & 1.20   & 1.30   \\
    \hline
  \end{tabular}
  \label{tb:single-hardware}
\end{table*}

\begin{table*}[ht]
\caption{Compare computation cost and performance of multi-hardware search and single-hardware search. One unit of search cost is $\sim$90 hours of Cloud TPU v2-32 usage.}
\vspace{-3mm}
\centering
  \begin{tabular}{ l | c | c | c c c c c | c c }
    \hline
      & Search & Accu & \multicolumn{2}{c}{CPU} & & & & \multicolumn{2}{c}{MN-Norm}  \\
     \multirow{-2}{*}{Model} & Cost & (\%) & float & uint8 & \multirow{-2}{*}{GPU} & \multirow{-2}{*}{DSP} & \multirow{-2}{*}{EdgeTPU} & avg & max \\ 
      \hline
    Single-Hardware&5$\times$&76.2    & 38.6   &    13.9& 5.85   & 3.71   & 2.55   &    1.13&    1.21\\
    Multi-Hardware &1$\times$&75.8    &    31.0&    13.9&    5.40& 3.81   &    2.40&    1.06& 1.25   \\
    \hline
  \end{tabular}
  \label{tb:single-multi}
  \vspace{-2mm}
\end{table*}

Figure \ref{fig:model} shows the visualization of the discovered multi-hardware models. The number inside each grey box is the output filter size (number of channels) for that stage, which is applied to each of the colored blocks inside. For example, in Multi-AVG model, second box from top, `dwbottleneck 3x3 / 96.0' indicates an inverted bottleneck block where the kernel size of depthwise conv is 3x3, the filter size of the expanded layer is 96 and the output filter size is 32. `skip' denotes an identity operation. The $\times$2 strides on resolution are taken at the same places as the centered model MobilenetV3: at the beginning of 1st, 2nd, 3rd, 4th, 6th stages respectively.

Both of the multi-hardware models have light early layers and heavy later layers, 5x5 kernels also appear later in the network. This indicates that multi-hardware models tend to move the computation to later layers where accelerators may gain more computation advantage. In addition, the observation that fused inverted bottlenecks are not chosen for multi-hardware models indicates that operations only effective on a small subset of examined hardware are not preferable for multi-hardware optimization.

\subsection{Multi-Hardware Search v.s. Single-Hardware Search}

To show the effectiveness of multi-hardware search comparing with single-hardware search, we conduct single-hardware search, by using reward function in equation (\ref{eq:single_r}), for each optimized hardware on the same multi-hardware search space with the multi-hardware search. The results are shown in Table \ref{tb:single-hardware}. Note that we choose to compare the direct search results without scaling because that represents the best performance from each search. The following comparisons will implicitly consider the minor difference on accuracy.

As expected, the best performance on CPU uint8, GPU and DSP was obtained from searching for corresponding hardware. Since the multi-hardware search space does not contain SE and h-swish which are particularly effectively on CPU float, Single-CPU float model searched on this search space only gives sub-optimal performance. Single-DSP model gives similar or even better results on EdgeTPU than Single-EdgeTPU model does, which may due to the high correlation between DSP and EdgeTPU. By checking the overall latency metrics, Single-CPU uint8 model (highlighted) gives the best results in all single-hardware models.

Taking the best single-hardware search results and comparing with the multi-hardware model (we take Multi-AVG here as they have similar accuracy) in Table \ref{tb:single-multi}, we can see that the best single-hardware model performs on par, or even slightly better than multi-hardware model on normalized max metric. However, which hardware would give the best model is unknown until we get all single-hardware models. Therefore, though single-hardware search might get slightly better results than multi-hardware search, its computation cost is $N\times$ of that needed for multi-hardware search, which scales linearly with the number of hardware one wants to optimize on.

\section{Conclusions}
\label{sec:conclude}

In this paper, we introduced an important but large ignored factor in hardware-aware neural architecture designs for applications that may be deployed on multiple hardware: model deployment cost. Taking this factor into consideration, we proposed a solution that minimized deployment cost, as well as development cost, while still achieving reasonably good performance across wide variety of hardware. Specifically, the concept of multi-hardware search space that is compatible with all examined hardware has been introduced, as well as the normalized average and max metrics to compare models' overall performance among multiple hardware. The multi-hardware models found in our experiments give SOTA performance on a majority of the examined hardware, as well as closely correlated un-searched hardware. Comparing with single-hardware searches which have to be applied on each target hardware separately, multi-hardware search gives comparable overall performance in a single search/train session.

\vspace{2mm}
\noindent \textbf{Acknowledgements}: We would like to thank Mark Sandler, Jaeyoun Kim, Jiahui Yu, Mingxing Tan, Ruoming Pang, Quoc V. Le, Hartwig Adam for helpful feedback and discussion; Cheng-Ming Chiang, Guan-Yu Chen, Koan-Sin Tan, Yu-Chieh Lin from MediaTek for useful guidance on MediaTek benchmarks; and QCT (Qualcomm CDMA Technologies) AI SW team for feedback on optimization.

{\small
\Urlmuskip=0mu plus 1mu\relax
\bibliographystyle{ieee_fullname}
\bibliography{egbib}
}

\end{document}